\documentclass[a4paper,fleqn]{cas-dc}

\usepackage[round,authoryear]{natbib}
\usepackage{amsmath}
\usepackage[linesnumbered, ruled]{algorithm2e}
\SetKwRepeat{Do}{do}{while}%
\usepackage{soul}
\usepackage{bm}
\usepackage{color, xcolor}
\soulregister\cite7
\soulregister\citep7
\soulregister\ref7

\usepackage[caption=false,font=normalsize,labelfont=sf,textfont=sf]{subfig}
\usepackage{graphicx}

\usepackage{multirow}

\def\tsc#1{\csdef{#1}{\textsc{\lowercase{#1}}\xspace}}
\tsc{WGM}
\tsc{QE}

\begin{document}
\let\WriteBookmarks\relax
\def\floatpagepagefraction{1}
\def\textpagefraction{.001}

\shortauthors{Gengyue Han et~al.}
\shorttitle{}

\title [mode = title]{CycLight: learning traffic signal cooperation with a cycle-level strategy}

\author[1,2,3]{Gengyue Han}[style=chinese]
\fnmark[1]
\ead{gyhan@seu.edu.cn}

\author[1,2,3]{Xiaohan Liu}[style=chinese]
\fnmark[1]
\ead{lxh_fy97@163.com}

\author[1,2,3]{Xianyue Peng}[style=chinese]
\ead{xypeng@seu.edu.cn}

\author[1,2,3]{Hao Wang}[style=chinese]
\cormark[1]
\ead{haowang@seu.edu.cn}

\author[1,2,3]{Yu Han}[style=chinese]
\ead{yuhan@seu.edu.cn}

\affiliation[1]{organization={Jiangsu Key Laboratory of Urban ITS, Southeast University},
  city={Nanjing},
  country={China}}

\affiliation[2]{organization={Jiangsu Province Collaborative Innovation Center of Modern Urban Traffic Technologies},
  city={Nanjing},
  country={China}}
\affiliation[3]{organization={School of Transportation, Southeast University},
  city={Nanjing},
  country={China}}

\cortext[1]{Corresponding author.}

\fntext[1]{The first two authors contributed equally to this paper.}

\begin{abstract}
  This study introduces CycLight, a novel cycle-level deep reinforcement learning (RL) approach for network-level adaptive traffic signal control (NATSC) systems. Unlike most traditional RL-based traffic controllers that focus on step-by-step decision making, CycLight adopts a cycle-level strategy, optimizing cycle length and splits simultaneously using Parameterized Deep Q-Networks (PDQN) algorithm. This cycle-level approach effectively reduces the computational burden associated with frequent data communication, meanwhile enhancing the practicality and safety of real-world applications. A decentralized framework is formulated for multi-agent cooperation, while attention mechanism is integrated to accurately assess the impact of the surroundings on the current intersection. CycLight is tested in a large synthetic traffic grid using the microscopic traffic simulation tool, SUMO. Experimental results not only demonstrate the superiority of CycLight over other state-of-the-art approaches but also showcase its robustness against information transmission delays.
\end{abstract}

\maketitle

\section{Introduction}
\label{S1}

Adaptive traffic signal control (ATSC) aims to mitigate potential congestion and enhance traffic efficiency in urban road networks, by dynamically adjusting signal timing according to real-time traffic conditions \citep{Chu2019}. ATSC approaches have not only been extensively investigated in academic research, but have also gained wide recognition for their robustness and effectiveness in practical implementation across numerous cities worldwide. Classic strategies, like SCOOT \citep{Hunt1978} and SCAT \citep{Luk1983}, employ dynamic traffic coordination, leveraging vehicle count detectors to optimize signal cycles, splits, and offsets. However, due to the highly dynamic nature of traffic operation, implementing efficient ATSC requires intricate adjustments to accommodate diverse traffic conditions. Since the 1990s, a variety of techniques have been proposed to optimize the ATSC system, encompassing model-based methods \citep{Daganzo1995,Mohajerpoor2019}, max pressure-based method \citep{Varaiya2013}, simulation-based optimization \citep{Chong2018,Osorio2017}, and data-driven approaches \citep{Li2016,Wiering2000}.

In recent years, the reinforcement learning (RL)-based technique, as a popular data-driven method, has gained increasing attention due to its capability of online traffic signal optimization without prior knowledge about the given environment \citep{RichardS.Sutton2018}. The RL-based controllers can learn from the interactions with the environment via trial and error without relying on pre-defined rules which are often used in conventional methods \citep{Han2022}. Initially, RL applications to ATSC were investigated in an isolated intersection \citep{Thorpe1996,Abdulhai2003}. The impressive results achieved by standard RL demonstrated the superiority of this data-driven approach over conventional methods. Nevertheless, computational burdens pose challenges to both training efficiency and control effectiveness with the scaling of the traffic network \citep{Haydari2020}. Therefore, several deep RL-based ATSC methods embedded with deep neural networks (DNN) were proposed, enabling RL agents to effectively recognize and process high-dimensional states, as well as facilitate the approximation of value functions \citep{Mnih2015}.

Even though DNN has significantly advanced the development of RL, training a single agent for network-level adaptive traffic signal control (NATSC) remains infeasible. Recent studies have explored the application of multi-agent reinforcement learning (MARL) \citep{LucianBusoniu2010} in NATSC. MARL emphasizes collaboration among agents, which can be categorized into centralized and decentralized settings according to various information structures \citep{ZhangK.YangZ.Basar2019}. Representative centralized methods include QMIX \citep{Rashid2018} and VDN \citep{Sunehag2012} were already utilized in NATSC, where a global single agent controls multiple local agents \citep{Lee2020}. Building upon this approach, subsequent research \citep{Wei2019} condensed the global scope into a smaller neighborhood. Graph convolution networks (GCN) and attention mechanism were introduced to facilitate coordination. However, centralized methods suffer from high latency, increased failure rates in practice, and the loss of topological information within the traffic network. Further, the joint action space grows exponentially as the complexity of traffic networks increases. Therefore, it is efficient and natural to formulate a decentralized NATSC system, where each intersection is controlled by a local RL agent, upon local observation and limited communication \citep{Chu2019}. The decentralized architecture is believed to be scalable, as the training and inference can be performed in parallel across intersections \citep{Liu2021}. Notably, as the cooperation in the decentralized system is achieved by information sharing among the neighboring agents, the delay of information transmission between intersections must be taken into account.

Another challenge that arises in the RL-based NATSC system is its practicability. Within the framework of RL, there are three commonly used methods for action selection at a single intersection \citep{Haydari2020}. The first method involves choosing a green phase from all possible phase sets, which is the most frequently employed \citep{Li2016,Pol2016}. The second method is a binary action selection that allows for either maintaining the current phase or transitioning to the next one \citep{Lin2018,Wei2018}. Finally, the less common method involves updating the phase duration according to a predefined length \citep{Casas2017,Yazdani2023}. Although these methods offer excellent control and timely decision-making, they impose significant computational burden due to high-frequency data communication in real-world applications \citep{Shabestary2020}. Moreover, disordered phase switching can greatly impact the driving experience and increase the risk of traffic accidents. Different from the step-by-step control strategy aforementioned, the cycle-level strategy is more preferred considering the issue of practicability.

However, there are seldom papers on cycle-level RL-based ATSC, primarily due to the intricate and extensive action space involved \citep{Shabestary2020}. Cycle-level controllers encounter a continuous action space where the duration of both the cycle and each individual phase can vary. Even if time is discretized, the action space expands significantly as the number of feasible phases increases. It is worth noting that commonly used MARL approaches typically generate either discrete or continuous actions, whereas cycle-level ATSC requires the simultaneous adjustment of discrete cycle length and continuous splits. Several studies compromise the optimization effectiveness, adjusting the splits under a constant cycle length \citep{Chin2011,Abdoos2021,Abdoos2014}. Besides, \cite{Wang2018} optimized the cycle length based on model-driven methods, while the splits were determined by RL agents. However, the relationship between the cycle and splits was not well established, making separate optimization susceptible to converging to local optimum. \cite{Shabestary2020} adopts RL agents to produce continuous actions, where each action indicated the duration of a specific phase. This method imposes no limit on the cycle length, resulting in large exploration domains that are hard to learn.

In this paper, we propose a novel cycle-level RL-based approach for NATSC, namely CycLight. The proposed approach leverages Parameterized Deep Q-Networks (PDQN) algorithm, performing discrete-continuous hybrid actions to optimize cycle length and splits simultaneously. Specifically, the cycle length is decided by discrete action, while the splits are represented as continuous parameters. During the joint evolution of both the discrete action and continuous parameters, CycLight is expected to find the optimal cycle length as well as avoid exhaustive search over continuous splits. To the best of our knowledge, this is the first paper that adopts MARL with discrete-continuous hybrid action space for cycle-level NATSC. To facilitate cooperation among intersections, a decentralized framework is formulated, where each local agent at intersection cooperates with others through information sharing. Moreover, the attention mechanism is embedded to correct the influence weight of surroundings on the current intersection. The proposed approach is tested in a 5*5 synthetic traffic grid using microscope simulation tool, SUMO \citep{Lopez2018}. Various traffic demand scenarios are designed to simulate different traffic distributions and flow rates. The proposed approach is compared with existing RL-based NATSC strategies as well as other state-of-art approaches. Additionally, we assess the robustness of CycLight by considering the delay of information transmission among intersections.

The rest of this paper is structured as follows. Section \ref{S2} provides an overview of the preliminary concepts. Section \ref{S3} introduces the framework of CycLight method. In Section \ref{S4}, the results of simulation experiments are presented in detail. Finally, concluding remarks and future implications are discussed in Section \ref{S5}.

\section{Preliminary}
\label{S2}

\subsection{Problem formulation}
\label{S2_A}

Network-level road is a typical urban traffic scenario that encompasses numerous intersections spread across a large-scale area. It bears the brunt of the traffic volume and therefore requires effective coordination among all intersections to achieve smooth travel for most vehicles. By employing MARL, the NATSC problem can be treated as a Markov Decision Process (MDP) $\mathbf{M} = \{ \mathbb{S},\mathbb{A},\mathcal{P},\mathbb{R},\gamma \} $. Each intersection is controlled by a unique agent, which can interact with the environment and gain an optimal strategy of action decision. At each time step $t$, assuming the MDP is in state ${s_t} \in \mathbb{S}$, the agent selects an action ${a_t} \in \mathbb{A}$, subsequently observing an immediate reward $r\left( {{s_t},{a_t}} \right)$ and transiting to the next state ${s_{t + 1}}$. The transition function can be represented by $\mathcal{P}:\mathbb{S} \times \mathbb{A} \times \mathbb{S} \to \mathbb{R}$. The objective of the agent is to maximize the discounted cumulative reward function ${\mathbb{E}}\left[ {\sum\nolimits_{t = 0}^T {{\gamma ^t}{r_t}} } \right]$ by continuously exploring and exploiting, through constant interactions with the environment. In this paper, the problem requests to reduce the average waiting time of each vehicle and improve the throughputs of the whole network. 

\begin{figure*}[!ht]
  \centering
  \includegraphics[width=0.9\textwidth]{ 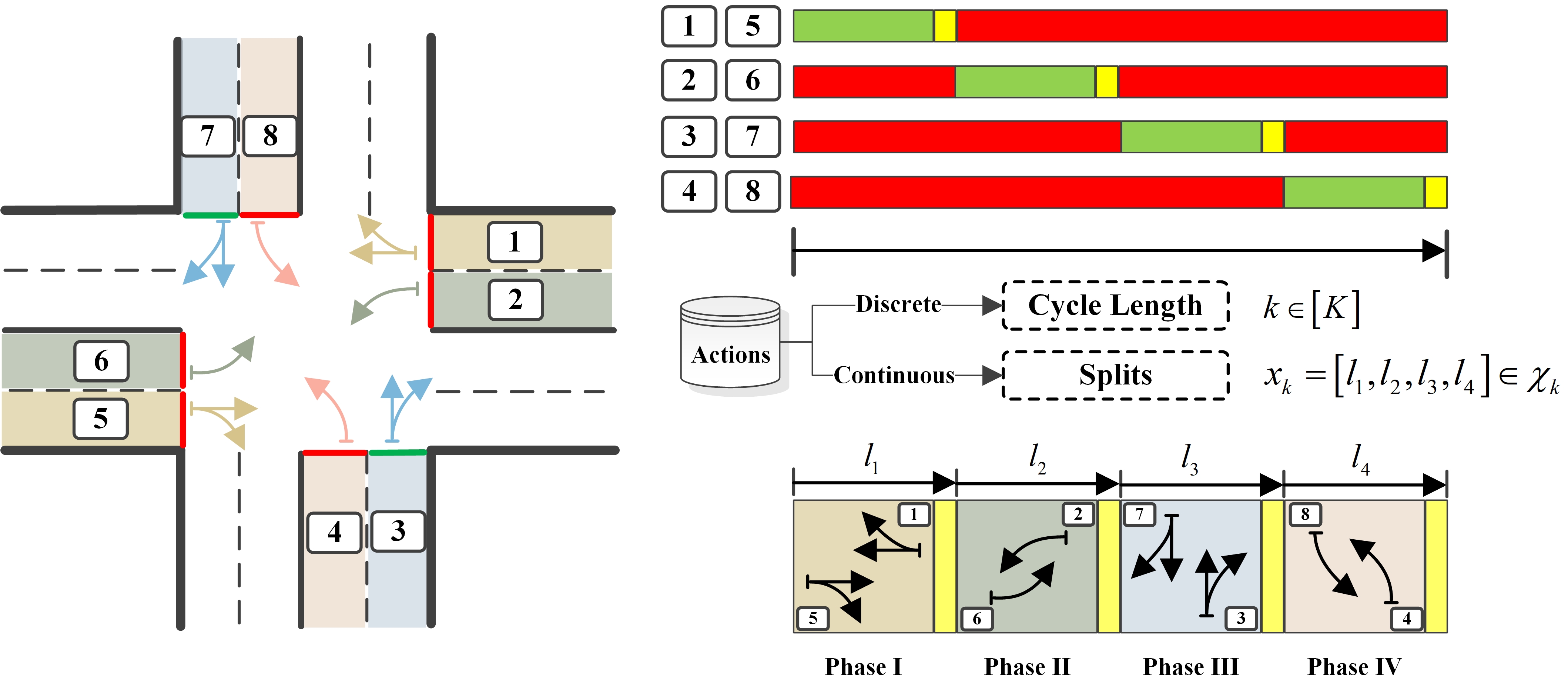}
  \caption{Illustration of a cycle-level TSC learning with discrete-continuous hybrid actions. The cycle length is decided by discrete action, while the splits are represented as continuous parameters.}
  \label{F1}
\end{figure*}

Notably, with the cycle-level strategy, the MDP with a parameterized action space is considered. As shown in Figure \ref{F1}, the agents performing discrete-continuous hybrid actions to optimize cycle length and splits simultaneously. 

\subsection{Parameterized Deep Q-Networks (PDQN)}
\label{S2_B}

PDQN, as one of the state-of-art RL methods, combines the spirits of both DQN (dealing with discrete action space) \citep{Mnih2013} and DDPG (dealing with continuous action space) \citep{Lillicrap2016} by seamlessly integrating them \citep{Xiong2018}. Accordingly, the Parameterized Action Markov Decision Process (PAMDP) \citep{Masson2016} is formulated as $ \mathbf{\tilde{M}} = \{ \mathbb{S},\mathbb{H},\mathcal{P},\mathbb{R},\gamma \} $. PAMDP is an extension of standard MDP with a discrete-continuous hybrid action space $\mathbb{H}$:
  \begin{equation}
  \label{eq1}
    \mathbb{H} = \{ (k,{x_k})|{x_k} \in {\chi _k}\;{\rm{for}}\;{\rm{all}}\;k \in [K]\},
  \end{equation} 
where $\left[ K \right] = \{ 1, \cdots ,K\} $ is the discrete action set; and ${\chi _k}$ is the corresponding continuous parameter set for each $k \in [K]$. In turn, we have state transition function $\mathcal{P}:\mathbb{S} \times \mathbb{H} \times \mathbb{S} \to \mathbb{R}$, reward function $\mathcal{R}:\mathbb{S} \times \mathbb{H} \to \mathbb{R} $, agent’s policy $\pi :\mathbb{S} \to \mathbb{H}$ and hybrid-action value function $Q\left( {s,k,{x_k}} \right)$ \citep{Li2022}. Then the Bellman equation is derived as:
  \begin{equation}
  \label{eq2}
    Q\left( {s,k,{x_k}} \right) = \mathop {\mathbb{E}}\limits_{r,{s^ - }} \left[ {r + \gamma \mathop {\max }\limits_{k' \in [K]} \mathop {\sup }\limits_{{x_k}^\prime  \in {\chi _k}} Q\left( {{s^ - },k',{x_k}^\prime } \right)} \right],
  \end{equation}
where ${s^ - }$ denotes the next state after taking the hybrid action $(k,{x_k})$. The cycle length and splits illustrated in Figure \ref{F1} are decided by the discrete $k$ and continuous ${x_k}$ from PDQN agent, respectively.  

\section{Method}
\label{S3}

This section offers a comprehensive overview of CycLight. To begin, PAMDP settings are specifically tailored for the cycle-level ATSC system. Subsequently, we present a decentralized control framework designed for NATSC.

\subsection{PAMDP formulations}
\label{S3_A}

A well-designed PAMDP is utilized to accurately capture key traffic flow features while minimizing computational load. In the case of a signalized intersection operating at cycle level, an RL agent engages with the environment to acquire real-time information and executes hybrid control actions at the end of each cycle. The cycle length and splits of the subsequent cycle are contingent upon the hybrid control action undertaken.

\subsubsection{State}
\label{S3_A_1}

In the conventional step-by-step control strategy mentioned in Section \ref{S1}, agents gather instantaneous observations from the environment as states and take actions at regular intervals, typically of a short time slot such as 5 or 10 seconds. Given the relatively stable traffic dynamics during brief intervals, the collected instantaneous observations are logically valid. Conversely, this notion does not hold true in the cycle-level strategy, as the control intervals are more widely spaced due to the longer duration of the cycle length compared to the shorter time slots. Consider a scenario where an RL controller performs actions at the end of the last phase and consequently receives the instantaneous observations at the same time. The lanes controlled by the last phase always exhibit reduced queues (or queues shorter than the average) since they have just been served, making it challenging for agents to accurately recognize the real-time traffic dynamics. Therefore, time-series data gathered throughout a complete cycle, rather than instantaneous observations, prove more appropriate for informing the cycle-level strategy. 

\begin{figure*}[!ht]
  \centering
  \subfloat[]{\includegraphics[width=0.33\textwidth]{ 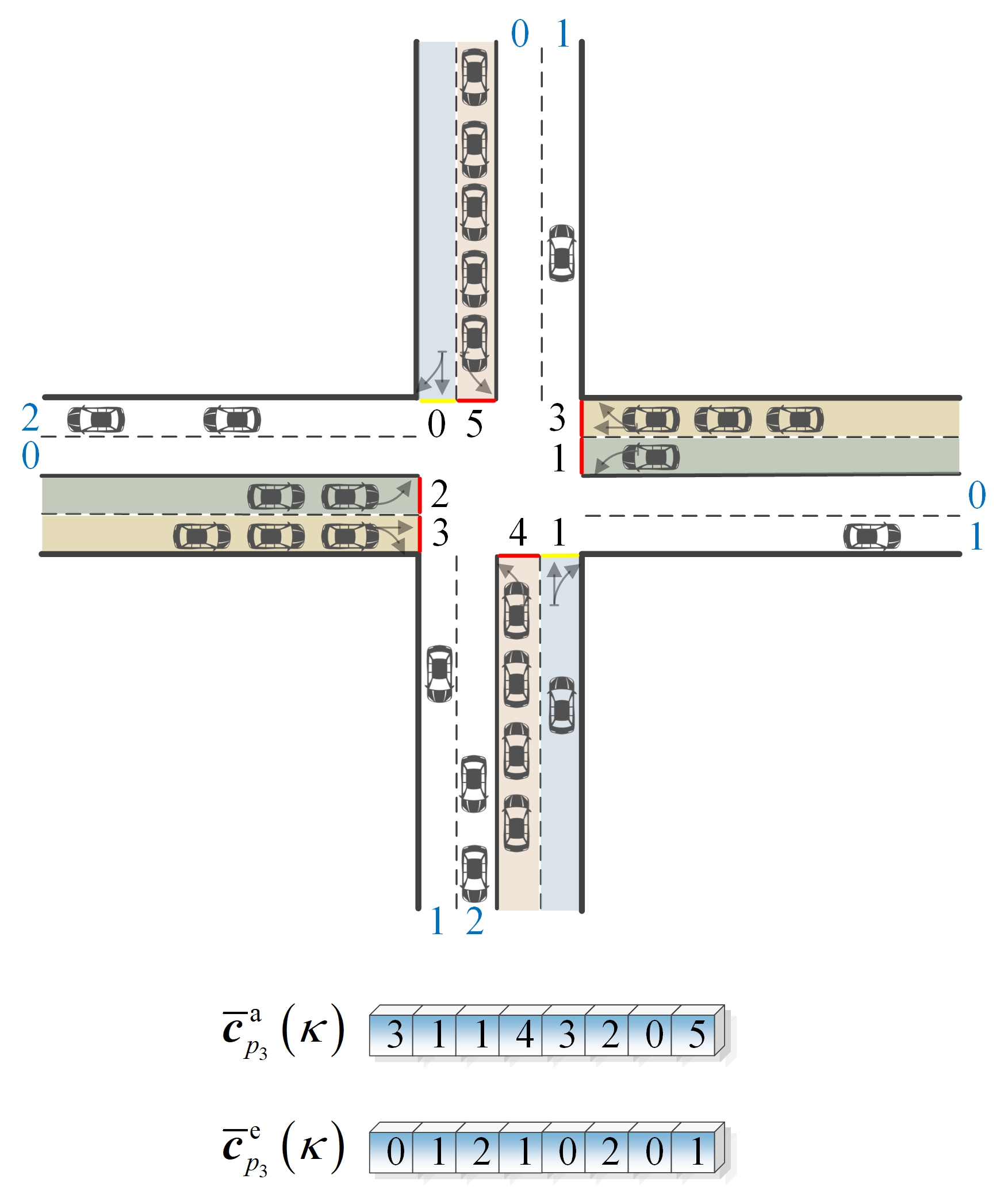}
  \label{F2_a}}
  \hfil
  \subfloat[]{\includegraphics[width=0.65\textwidth]{ 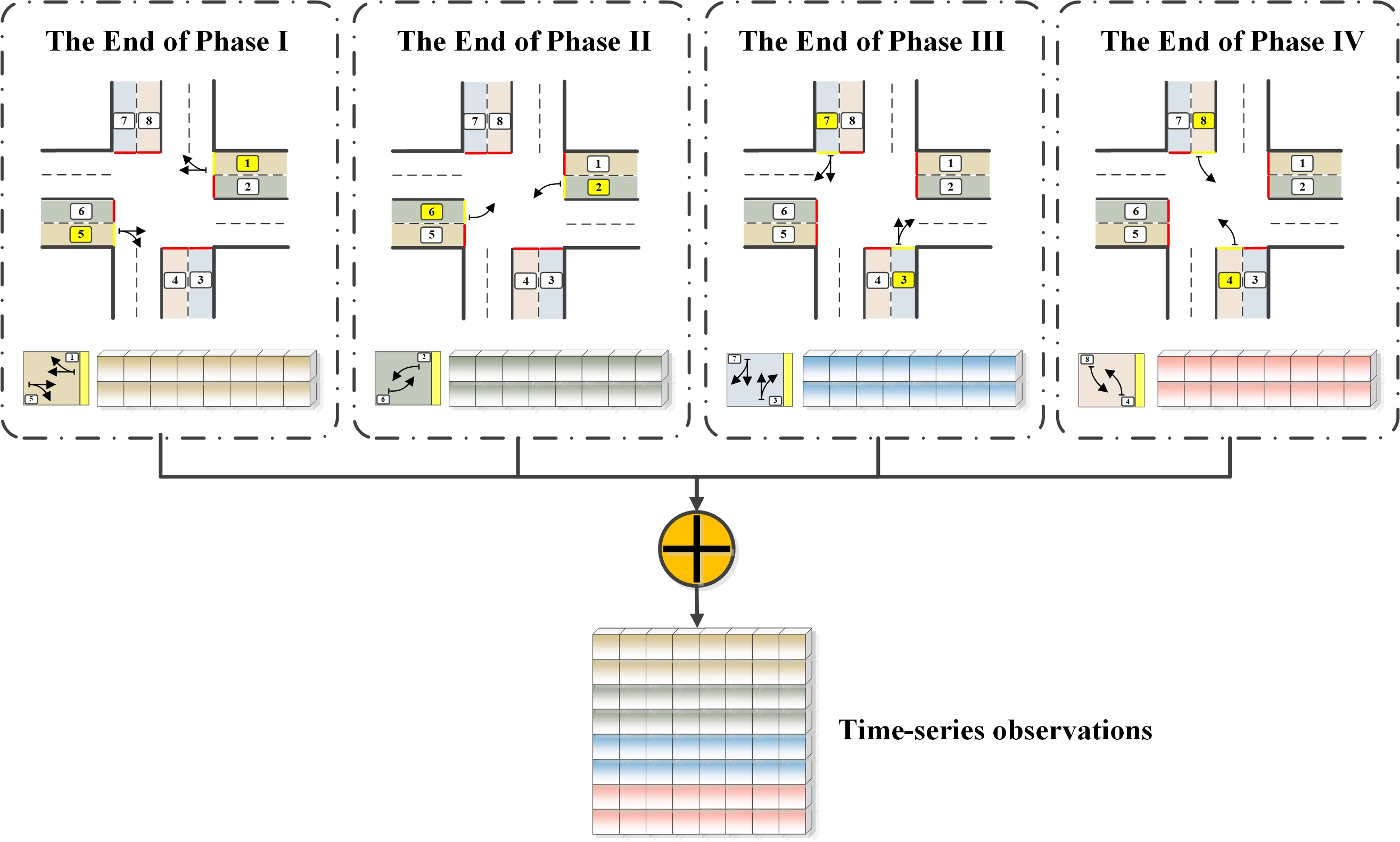}
  \label{F2_b}}
  \caption{The process of observed states gathering. (a) is an example of RL-based ATSC at a single intersection. The lane area sensors detect the vehicle counts on each approach and exit at the end of Phase III, and store them as $\bar c_p^{\rm{a}}\left( \kappa  \right)$ and $\bar c_p^{\rm{e}}\left( \kappa  \right)$, respectively. (b) shows the formulation of time-series observations, which are gathered throughout a complete cycle.
     }
  \label{F2}
  \end{figure*}

Figure \ref{F2} displays the formulation of traffic states using time-series data. As is shown in Figure \ref{F2} (a), the lane area sensors detect the vehicle counts on each approach and exit, within a distance of 100 meters from the corresponding stop lines. The vectors $\bar c_p^{\rm{a}}\left( \kappa  \right)$ and $\bar c_p^{\rm{e}}\left( \kappa  \right)$ are introduced to store the detected vehicle counts at the end of each phase. Integrating time-series data throughout a complete cycle, the time-series observations are formulated. Specifically, the vectors $\bar c_p^{\rm{a}}\left( \kappa  \right)$ and $\bar c_p^{\rm{e}}\left( \kappa  \right)$ are collected at the end of each phase within a cycle, and then the vectors are concatenated to form a time-series observation sequence at the end of the cycle. Subsequently, the time-series observation is fed into the RL agent for further processing. With access to the traffic conditions during the pivotal moments of phase changes, the agent can employ interpolation to approximate the traffic conditions at any intermediate point. The local time-series state of cycle $\kappa $ is represented as:
  \begin{equation}
  \label{eq3}
    {s^{{\rm{local}}}}\left( \kappa  \right) = \mathop  \cup \limits_{i = 1}^4 \left[ {\bar c_{{p_i}}^{\rm{a}}\left( \kappa  \right),\bar c_{{p_i}}^{\rm{e}}\left( \kappa  \right)} \right],
  \end{equation}  
where ${p_i}$ denotes the phase $i$. A detailed depiction of this process can be found in Figure \ref{F2} (b).

\subsubsection{Reward}
\label{S3_A_2}

In the ATSC system, rewards are designed to minimize the average delay of vehicles and enhance the efficiency of traffic flow at intersections. In our proposed cycle-level control strategy, each local reward is calculated based on the average waiting time experienced by vehicles during the previous cycle. To achieve this, the sensors detect the stopped waiting (speed is below than 0.1m/s) time of vehicles within a cycle, and the average waiting time at the intersection per vehicle is quantified as follows:
  \begin{equation}
  \label{eq4}
    w\left( \kappa  \right) = \frac{1}{V}\sum\limits_{v = 1}^V {\varpi \left( v \right)},
  \end{equation}
where $\varpi \left( v \right)$ denotes the waiting time of vehicle $v$ during the previous cycle; and $V$ is the total number of vehicles detected. Notably, longer cycle time inherently leads to increased cumulative delays within a cycle, whereas excessively short cycle lengths lead to decreased intersection capacity due to the additional loss time introduced by frequent phase shifts. The release of vehicles in the current cycle that exceeds the intersection capacity results in residual vehicles waiting in front of the stop line, leading to the formation of secondary queues. An optimal cycle length should strike a balance to ensure both lower cumulative delays and higher intersection capacity. Therefore, a penalty item $\Gamma \left( k \right) = {N^{S,T}} \cdot {\lambda _p} \left({\lambda _p} > 0\right)$ is introduced to penalize the number of vehicles ${N^{S,T}}$ in the secondary queues or even triple queues within the cycle. The reward function is given:
  \begin{equation}
  \label{eq5}
    R\left( \kappa  \right) =  - w\left( \kappa  \right) - \Gamma \left( \kappa  \right).
  \end{equation}
  
\subsubsection{Action}
\label{S3_A_3}
With PDQN, discrete-continuous hybrid actions are produced to determine the cycle length and splits, respectively. The cycle length ranges from 60 to 120 seconds, which is discretized by 12 seconds. Therefore, the discrete set $\left[ K \right]$ is defined as $\left[ K \right] = [60,72,84,96,108,120]$. The splits are controlled by continuous action ${x_k}$, where ${x_k} = \left[ {{l_1},{l_2},{l_3},{l_4}} \right]$. ${l_i}$ denotes the green scale factor of the phase $i$, and $\sum\nolimits_{i = 1}^4 {{l_i} = 1} $. Based on the cycle length $k$ and green scale factor ${l_i}$, the effective green time of each phase ${L_i}$ is derived as:
  \begin{equation}
  \label{eq6}
    {L_i} = \left( {k - \sum\nolimits_{i = 1}^4 {g_i^{\min }}  - \sum\nolimits_{i = 1}^4 {y_i^L} } \right) \cdot {l_i} + g_i^{\min },
  \end{equation} 
where $g_i^{\min }$ is the minimum green time of phase $i$; $y_i^L$ is the yellow light duration for connection. The implementation of the minimum green time has elevated the lower threshold of control effectiveness, effectively preventing the occurrence of extremely undesirable situations. 

\subsection{The architecture of CycLight}
\label{S3_B}

\begin{figure*}[!ht]
  \centering
  \includegraphics[width=0.9\textwidth]{ 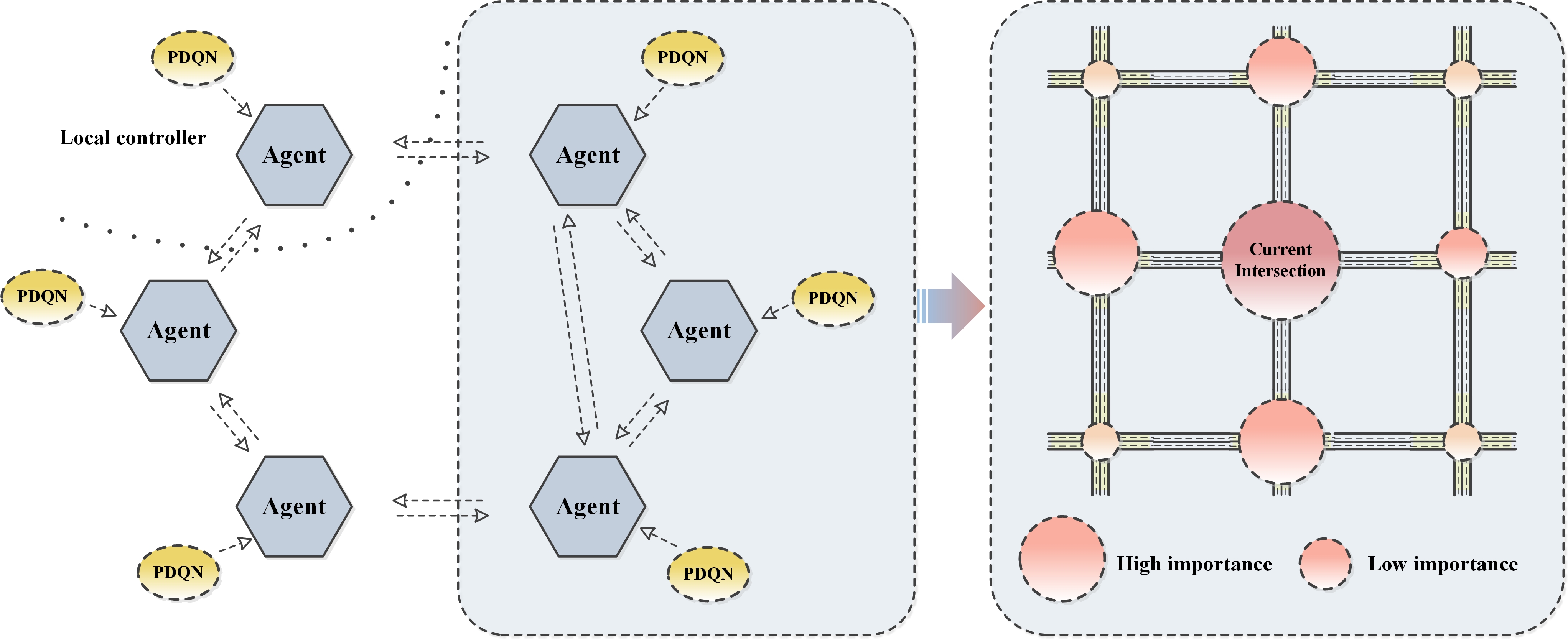}
  \caption{PDQN agents with decentralized setting. The current intersection attaches more attention to important neighbors during information sharing process.}
  \label{F4}
\end{figure*}

On the basis of well designed PAMDP, the multi-agent system can be established for NATSC. The proposed CycLight adopts decentralized framework, sharing important information with the neighbors. Besides, the attention mechanism is introduced to correct the influence weight of surroundings on the current intersection (Figure \ref{F4}). 

\begin{figure*}[!ht]
  \centering
  \includegraphics[width=0.9\textwidth]{ 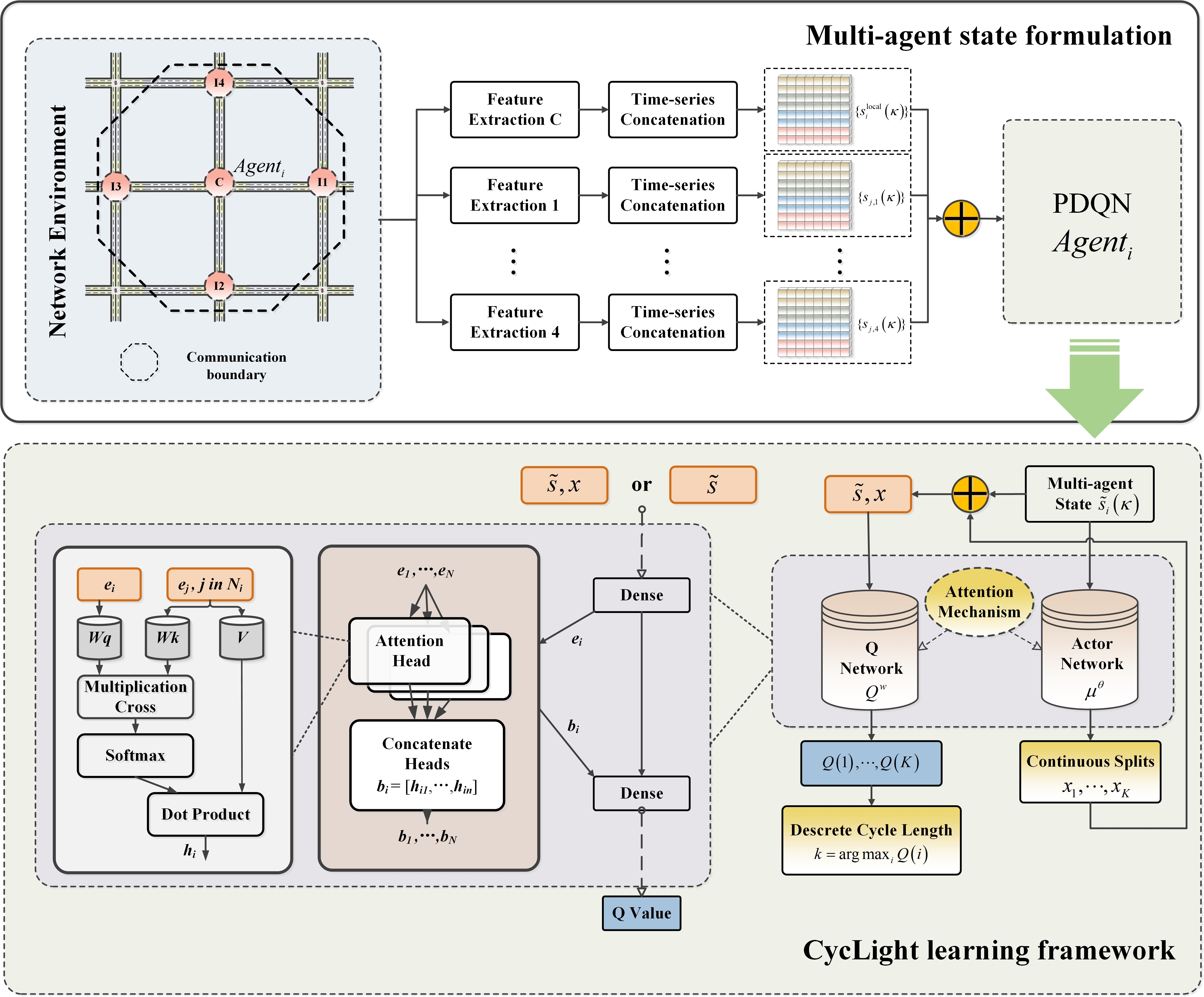}
  \caption{Detailed structure of CycLight.}
  \label{F5}
\end{figure*}

\subsubsection{Multi-agent extension}
\label{S3_B_1}

During the training process of CycLight, traffic features are detected and extracted by the agent, and subsequently the local time-series state $s_i^{{\rm{local}}}\left( \kappa  \right)$ defined in Eq.(\ref{eq3}) can be derived. As depicted in the top-left of Figure \ref{F5}, communication takes place among the current intersection and its adjacent neighbors. The neighbor set of the agent $i$ is defined as ${N_i}$. In most existing studies, the shared information typically comprises the local states of each neighbor agent. In our settings, some critical real-time control scheme and timelines are also incorporated, since the cycle-level TSC are inherently asynchronous. Current agent is supposed to be informed of the real-time phase rolling of the neighbor agents. Therefore, the multi-agent information-sharing vector ${m_i}\left( \kappa  \right)$ can be defined as:
  \begin{equation}
  \label{eq7}
   {m_i}\left( \kappa  \right){\rm{ = }}\left[ {s_j^{{\rm{local}}}\left( \kappa  \right),{\delta _j}\left( \kappa  \right)} \right],j \in {N_i},
  \end{equation}
where ${\delta _j}\left( \kappa  \right) = \left[ {{k_j},{x_{{k_j}}},{\rho _j}} \right]$; ${\rho _j}$ denoted the remaining time of the current cycle for agent $j$. Consequently, the state and reward in the mode of multi-agent extensions can be written as:
  \begin{equation}
  \label{eq8}
   {\tilde s_i}\left( \kappa  \right) = s_i^{{\rm{local}}}\left( \kappa  \right) \cup {m_i}\left( \kappa  \right),
  \end{equation}
  \begin{equation}
  \label{eq9}
   {\tilde R_i}\left( \kappa  \right) = \frac{1}{{\left| {{N_i}} \right| + 1}}\left[ {{R_i}\left( \kappa  \right) + \sum\limits_{j \in {N_i}} {\varphi {R_j}\left( \kappa  \right)} } \right],
  \end{equation}  
where $\left| {{N_i}} \right|$ is the total number of elements in the set ${N_i}$; while $\varphi  \in \left( {0,1} \right)$ is a constant discount factor.

\subsubsection{Updating process of CycLight}
\label{S3_B_2}

The updating process of CycLight is described in detail at the bottom of Figure \ref{F5}. PDQN agent employs the deterministic actor network from DDPG to generate continuous splits, while the Q-network from DQN is utilized to obtain discrete cycle length. To be specific, the observed states are fed into the actor network to solve $x_k^* = \arg {\sup _{{x_k} \in {\chi _k}}}Q\left( {s,k,{x_k}} \right)$ for each $k \in \left[ K \right]$, and then the largest $Q\left( {s,k,x_k^*} \right)$ is selected. The Q-network ${Q^w}$ with network weights $w$ is employed to approximate $Q\left( {s,k,{x_k}} \right)$. Additionally, we approximate ${x_k}\left( s \right)$ with the deterministic actor network ${\mu ^\theta }$, where $\theta $ denotes its network weights. Assuming $w$ is fixed, $\theta $ can be optimized based on the assumption:
  \begin{equation}
  \label{eq10}
   {Q^w}\left( {s,k,{\mu _\theta }\left( s \right)} \right) \approx {\sup _{{x_k} \in {\chi _k}}}Q\left( {s,k,{x_k}} \right)\;\;\;{\rm{for}}\;{\rm{each}}\;k \in \left[ K \right].
  \end{equation}
 $w$ in the Q-network could be estimated by the mean-squared Bellman error via gradient descent. In the $\kappa-{\rm{th}}$ control step, the n-step target $y\left( \kappa  \right)$ is derived as:
  \begin{equation}
  \label{eq11}
   y\left( \kappa  \right) = \sum\limits_{i = 0}^{n - 1} {{\gamma ^i}\tilde R\left( {\kappa  + i} \right)}  + {\gamma ^n}\mathop {\max }\limits_{k \in [K]} {Q^w}\left( {\tilde s\left( {\kappa  + n} \right),k,{\mu ^\theta }\left( {\tilde s\left( {\kappa  + n} \right)} \right)} \right).
  \end{equation}
 
The loss functions of the Q-network and actor network are displayed as follows:
  \begin{equation}
  \label{eq12}
   {\ell ^w}\left( \kappa  \right) = \frac{1}{2}{\left[ {{Q^w}\left( {\tilde s\left( \kappa  \right),k\left( \kappa  \right),{x_k}\left( \kappa  \right)} \right) - y\left( \kappa  \right)} \right]^2},
  \end{equation}
  \begin{equation}
  \label{eq13}
   {\ell ^\theta }\left( \kappa  \right) =  - \sum\limits_{k = 1}^K {{Q^w}\left( {\tilde s\left( \kappa  \right),k,{\mu ^\theta }\left( {s\left( \kappa  \right)} \right)} \right)} .
  \end{equation}
 
\subsubsection{Attention mechanism embedding}
\label{S3_B_3}

Notably, given the large dimension of the multi-agent state, there is a risk that the current agent may be overwhelmed by the complex inputs. To address this issue, an attention mechanism has been incorporated into both the Q-network and actor network, as depicted in Figure \ref{F5}. This mechanism enhances the accuracy of approximation by the DNNs. Specifically, the outputs from the Q-network for each agent can be reformulated as follows:
  \begin{equation}
  \label{eq14}
   Q_i^w\left( {{{\tilde s}_i},{k_i},{x_{{k_i}}}} \right) = \mathcal{F}\left( {{e_i},{b_i}} \right),
  \end{equation}
where $\mathcal{F}$ is a three-layer dense perceptron; and ${e_i}$ denotes the state-action factor obtained by passing through a single-layer perceptron $\mathcal{D}$ with the activation function $leaky ReLU$, i.e. ${e_i} = \mathcal{D}\left( {s_i^{{\rm{local}}},{k_i},{x_{{k_i}}},{\rho _i}} \right)$. Moreover, the information from the neighboring agents is extracted using ${b_i}$, which is a vector that concatenates the weighted sum of the value function of each agent.

In order to estimate ${b_i}$, we employ the multiple attention heads approach proposed by \cite{Vaswani2017}. Each attention head is assigned its own set of parameters $\left[ {{W_k},{W_q},V} \right]$, which collectively generate an aggregated contribution factor ${h_{il}}$ from all neighboring agents to the target agent. Since each attention head can focus on different weighted combination, the contribution factors ${h_{il}}$ are concatenated into the vector ${b_i}$. Specifically, for a model with $n$ attention heads, the output ${b_i} = \left[ {{h_{i1}}, \ldots ,{h_{in}}} \right]$ can be expressed as the sum of the contribution factors ${h_{il}}$, which is derived as follows:
  \begin{equation}
  \label{eq15}
  {h_{il}} =\sum\nolimits_{j \in {N_i}} {\alpha _{ij}} { {V}{ {e}_j}} ,l \in \left[ {1, \ldots ,n} \right],
  \end{equation}
where the state-action factor ${e_j} = \mathcal{D}\left( {s_j^{{\rm{local}}},{k_j},{x_{{k_j}}},{\rho _j}} \right)$. Then ${e_j}$ is linearly transformed by a shared matrix $V$. In addition, ${\alpha _{ij}}$ corresponds to attention score, indicating the degree of similarity between ${e_i}$ and ${e_j}$. Following a similar approach to the differentiable key-value memory model \citep{Oh2016}, a query-key technique is employed. Specifically, ${ {W}_q}$ transforms ${ {e}_i}$ into a query, while ${ {W}_k}$ transforms ${ {e}_j}$ into a key. ${\upsilon _{ij}}= e_j^TW_k^T{W_q}{e_i}$ reflects the influence of the adjacent intersection $j$ on the current intersection $i$. The attention score is derived as:
  \begin{equation}
  \label{eq16}
  {\alpha _{ij}} = \frac{{\exp \left( {{\upsilon _{ij}}} \right)}}{{\sum\nolimits_{k \in  N_i} {\exp \left( {{\upsilon _{ik}}} \right)} }}.
  \end{equation}
 
Similarly, the outputs from the actor network can also be rewritten as:
  \begin{equation}
  \label{eq17}
   \mu _i^\theta \left( {{{\tilde s}_i}} \right) = \mathcal{F}'\left( {{{e'}_i},{{b'}_i}} \right).
  \end{equation}
 
It is worth mentioning that the attention mechanism serves as a supplementary tool in DNNs for approximating Q values. Its primary purpose is to adjust the influence weights of the surroundings on the current intersection in the multi-agent system. The embedding of attention mechanism is not contradictory to Eq.(\ref{eq12}) and (\ref{eq13}) in any sense.

Algorithm \ref{al1} describes the pseudo-code of CycLight. Given an environmental state, we obtain a hybrid action so as to interact with the environment. Then the collected transition sample is stored in the replay buffer, after which the policy learning is performed using the data sampled from $D$. At the end of learning, the converged parameters are saved, so that the hybrid optimal action set $\left( {{k^{\rm{*}}},x_k^*} \right)$ can be obtained according to:
  \begin{equation}
  \label{eq18}
   \left\{ \begin{array}{l}
x_k^* = {\mu ^{{\theta ^*}}}\left( {\tilde s} \right)\\
{k^*} = \arg {\max _{k \in [K]}}{Q^{{w^*}}}\left( {\tilde s,k,x_k^*} \right)
\end{array} \right..
  \end{equation}
 
  \begin{algorithm*}[!t]
    {\bf Input:} Step-size $\{ {\beta ^w},{\beta ^\theta }\} $, reward discount factor $\gamma $, exploration parameter $\varepsilon $, minibatch size $\left| B \right|$, a probability distribution $\xi $, time horizon per episode $T$, episode horizon $\Xi $, PAMDP parameters ${\lambda _p}$, $\varphi $, ${g^{\min }}$, $y_i^L$  \\
    {\bf Initialize} Q-network ${Q^w}$ and actor network ${\mu ^\theta }$ with random parameters $w$, $\theta $\\ 
    {\bf Initialize} attention head parameters $\left[ {{W_k},{W_q},V} \right]$\\ 
    {\bf Initialize} $E \leftarrow 0,t \leftarrow 0,\kappa  \leftarrow 0$\\
    {\bf Prepare} replay buffer $D = \phi $ \\
    \Repeat{reaching maximum total environment steps: $E = \Xi $}
    {
        \For{$t = 0,1,2, \ldots ,T$}
        {
          \If {Control condition}
          {
            Compute continuous actions ${x_k}\left( \kappa  \right) = {\mu ^\theta }\left( {\tilde s} \right),{x_k}\left( \kappa  \right) \in {\chi _k}$ with attention mechanism 
            \hfill $\triangleright$ see Eq. (\ref{eq17}) \\
             Select discrete actions based on $\varepsilon  - greedy$ policy with attention mechanism: 
             \hfill $\triangleright$ see Eq. (\ref{eq14}) \\
            $k\left( \kappa  \right) = \left\{ \begin{array}{l}{\rm{a}}\;{\rm{sample}}\;{\rm{from}}\;{\rm{distribution}}\;\xi \quad \quad \quad \quad \;\;{\rm{with}}\;{\rm{probability}}\;\varepsilon \\ 
            \arg {\max _{k \in [K]}}{Q^w}\left( {\tilde s\left( \kappa  \right),k,{x_k}\left( \kappa  \right)} \right)\quad \quad {\rm{with}}\;{\rm{probability}}\;1 - \varepsilon 
            \end{array} \right.$ \\
            Execute $\left( {k\left( \kappa  \right),{x_k}\left( \kappa  \right)} \right)$, observe $\tilde R\left( \kappa  \right)$ and transmit to the next state $\tilde s\left( {\kappa  + 1} \right)$  
            \hfill $\triangleright$ see Eq. (\ref{eq8}) and (\ref{eq9}) \\
            Store transition $\left[ {\tilde s\left( \kappa  \right),k\left( \kappa  \right),{x_k}\left( \kappa  \right),\tilde R\left( \kappa  \right),\tilde s\left( {\kappa  + 1} \right)} \right]$ into $D$ \\
            Sample a mini-batch of $B$ experience from $D$  \\
            Compute loss functions ${\ell ^w}\left( \kappa  \right)$ and ${\ell ^\theta }\left( \kappa  \right)$
            \hfill $\triangleright$ see Eq. (\ref{eq12}) and (\ref{eq13}) \\
            Update the weights by ${w_{\kappa  + 1}} \leftarrow {w_\kappa } - {\beta ^w}{\nabla _w}{\ell ^w}\left( \kappa  \right)$ and ${\theta _{\kappa  + 1}} \leftarrow {\theta _\kappa } - {\beta ^\theta }{\nabla _\theta }{\ell ^\theta }\left( \kappa  \right)$ \\
            $\kappa  \leftarrow \kappa  + 1$ \\
          }
        } 
        $E \leftarrow E + 1$\\       
    }
    \caption{CycLight}
  \label{al1}
  \end{algorithm*}

\section{Experiments}
\label{S4}

CycLight is evaluated in a large synthetic traffic grid simulated by SUMO. This section focuses on designing challenging and time-varying traffic environments to facilitate fair comparisons among different controllers.

\subsection{General setups}
\label{S4_A}

\begin{figure}[!ht]
  \centering
  \includegraphics[width=0.48\textwidth]{ 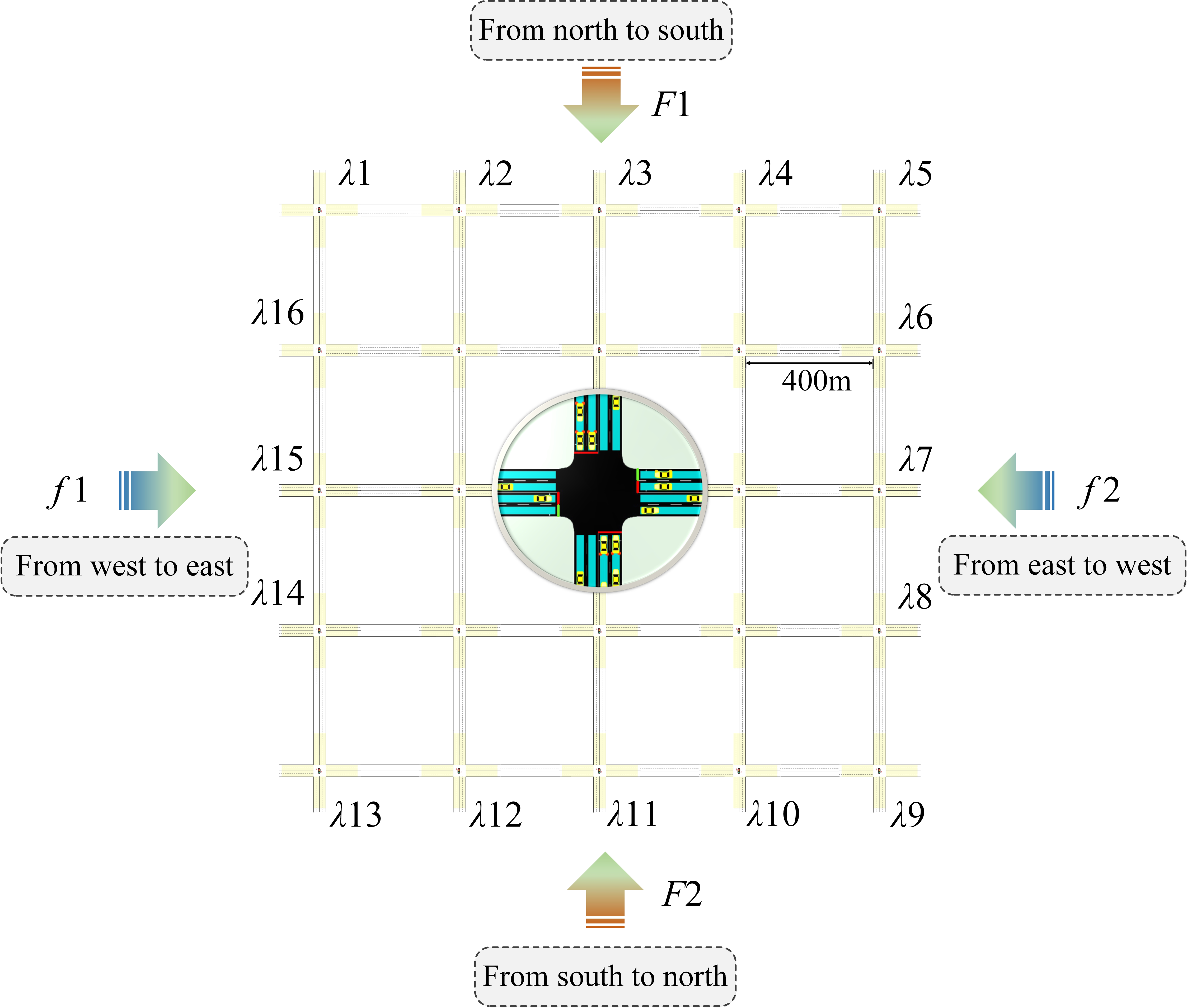}
  \caption{The synthetic traffic gird test-bed.}
  \label{F6}
\end{figure}

\begin{table}[!htbp]
  \begin{center}
  \caption{The demand upper bound for each flow group.}
  \label{tab1}
  \begin{tabular}{ c  c }
  \hline
   Flow group & Upper bound (pcu/h)\\
  \hline
  $F1$ & 300 \\
  $F2$ & 350 \\
  $f1$ & 200 \\
  $f2$ & 250 \\
  \hline
  \end{tabular}
  \end{center}
  \end{table}

\begin{table}[!h]
  \renewcommand\arraystretch{1.2}
  \begin{center}
  \caption{The setups of RL controller.}
  \label{tab2}
  \begin{tabular}{ c c c }
  \hline
  & Parameters & Value \\
  \hline
  &${\beta ^w}$ & $0.001$\\
  &${\beta ^\theta }$ & $0.001$\\
  RL&$\gamma$ & $0.99$\\
  hyper-parameters&$\left| B \right|$ & $128$\\
  &$\varepsilon $ & $0.05 \sim 1$\\
  &$n$ & $4$\\
  \cline{1-3}
  &$\lambda _p$ & $10$\\
  MDP&$\varepsilon $ & $0.9$\\
  parameters&${g^{\min }}$ & $12s$\\
  &$y_i^L$ & $3s$\\
  \hline
  \end{tabular}
  \end{center}
  \end{table}

Figure \ref{F6} presents the schematic diagram of the synthetic 5*5 traffic grid considered in this study. The traffic grid comprises two-lane arterial streets where the speed limit is set at 20 m/s. In order to create a challenging PAMDP, we simulate four groups of time-varying traffic flows with specific origin-destination (O-D) pairs. Within the first flow group, denoted as $F1$, there are five origins represented by $\{ {\lambda _1}, \cdot  \cdot  \cdot ,{\lambda _5}\} $. Each origin generates traffic flows towards one randomly selected destination from the set of destinations $\{ {\lambda _9}, \cdot  \cdot  \cdot ,{\lambda _{13}}\} $ in the opposite direction. The selection of destinations is determined by the episode random seed $\zeta \left( E \right)$, which remains constant throughout each episode $E$. Consequently, there are five possible routes for each flow group in every episode. A similar traffic flow generation paradigm is also adopted for flow groups $F2$, $f1$, and $f2$. The traffic demands for these flow groups are determined by multiplying the upper bounds specified in Table \ref{tab1} with a random factor $\sigma  \in \left( {0,1} \right)$. To capture various traffic dynamics, $\sigma$ is reset every 300 seconds. As a result, the time-varying traffic demand displays an uneven distribution, meaning that the traffic patterns within neighboring intersections of a targeted intersection may significantly differ. Hence, the superiority of the attention mechanism embedded algorithm becomes evident in handling such variations.

Besides, the time horizon per episode $T$ is set to 3000 seconds. The RL models are trained for $\Xi  = 700$ episodes with different random seeds. Several RL hyper-parameters and PAMDP parameters are displayed in Table \ref{tab2}:

\subsection{Testing scenarios}
\label{S4_B}

\subsubsection{State-of-art ATSC baselines}
\label{S4_B_1}

To evaluate the control performance of the proposed CycLight, several state-of-art ATSC controllers are selected as baselines, i.e. 1) Single PDQN, 2) MAADDPG, 3) Cycle-level BackPressure and 4) Adaptive Webster control strategies. 

{\bf Baseline 1:} With single PDQN, each intersection is controlled by an independent PDQN agent with no information sharing among the neighbors. Model parameters are updated according to the local states detected at local intersections.

{\bf Baseline 2:} Multi-Agent Attention Deep Deterministic Policy Gradient (MAADDPG) is the multi-agent extension of DDPG, designed to tackle multi-agent scenarios. It also incorporates attention mechanism into the critic network, further enhancing its capabilities of recognizing important surroundings. Since DDPG utilizes the deterministic policy that enables continuous action selection, the agent directly outputs the duration of green light for each phase, without being constrained by cycle length. To avoid exceedingly high traffic delay, a maximum limit of 60 seconds is imposed on the duration of each green phase.

{\bf Baseline 3:} The cycle-level BackPressure controller \citep{Le2015} determines the splits within a pre-defined fixed cycle length, taking into consideration the vehicle pressure of each phase. This controller is an adaptation of the well-regarded BackPressure scheme, which has gained widespread recognition in the field of NATSC.

{\bf Baseline 4:} The Adaptive Webster controller \citep{Genders2019} collects data over a specific time interval, denoted as $I$, and utilizes Webster's method to compute the cycle length and splits for the subsequent time interval. This adaptive approach essentially relies on the most recent $I$ interval to gather data, while assuming that the traffic demand will remain relatively stable during the upcoming interval. 

The four baselines mentioned above correspond to testing {\bf Scenario A.1-4}, respectively.

\subsubsection{Advance control considering information transmission delay}
\label{S4_B_2}

Given that cooperation among agents relies on information sharing, it is crucial to consider the delay in information transmission between intersections. In our proposed CycLight system, RL agents collect the multi-agent state at the end of each cycle and subsequently execute control actions based on the information. This interaction process can be easily replicated in simulators. However, in real-world scenarios, there are information transmission delays, especially during periods of communication congestion. Consequently, agents may encounter challenges in executing actions promptly. To tackle this issue, we try to perform advance control. Specifically, agents gather the environmental states approximately 5 seconds prior to the end of each cycle, allowing them to devise appropriate phase plans for the subsequent cycle. The advance control strategy, denoted as Advance CycLight, is referred to as {\bf Scenario B}.

\subsection{Performance evaluation}
\label{S4_C}

\subsubsection{Training results}
\label{S4_C_1}

\begin{figure}[!ht]
  \centering
  \includegraphics[width=0.45\textwidth]{ 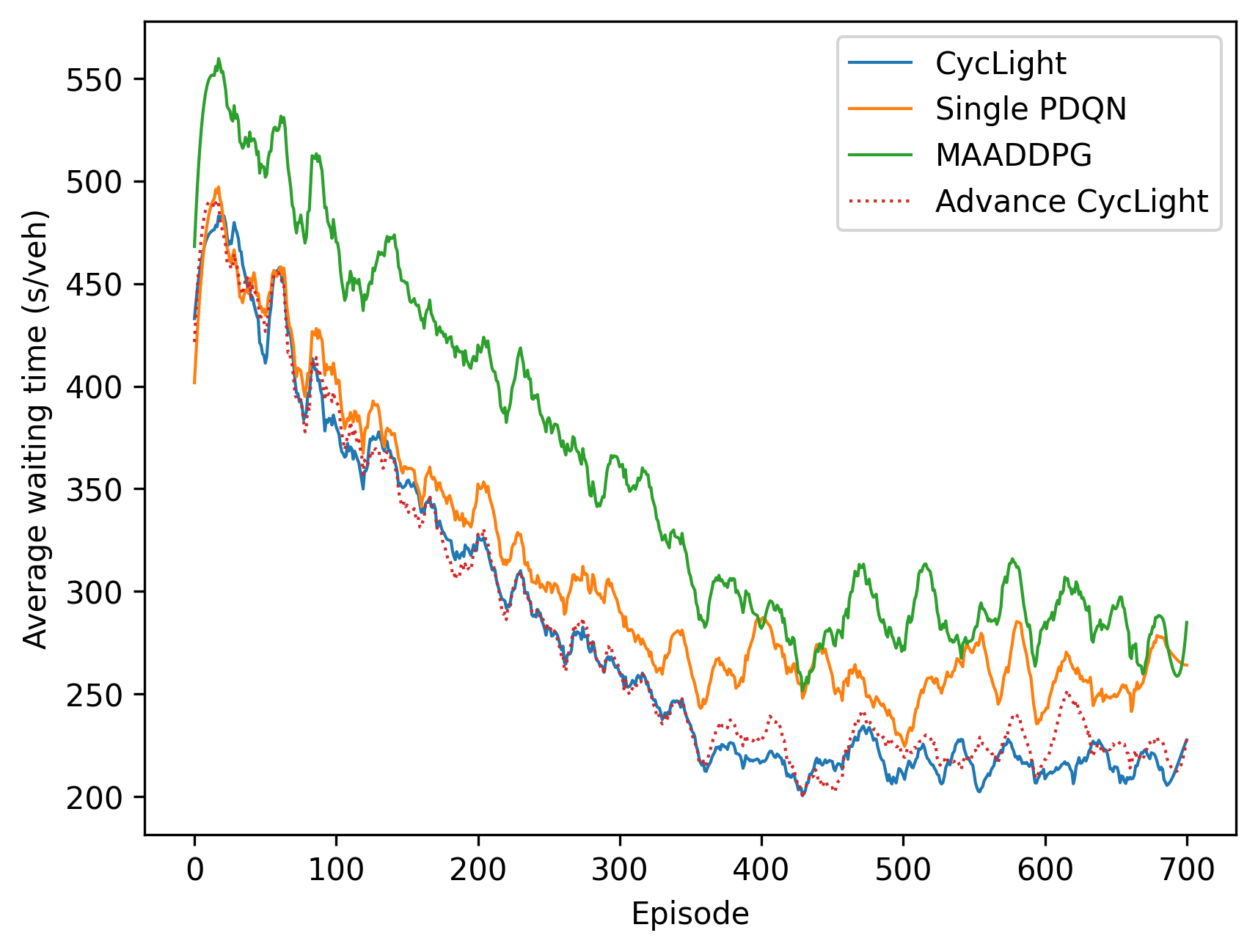}
  \caption{Training curve of each RL-based controller.}
  \label{F7}
\end{figure}

Figure \ref{F7} illustrates the training curve of each RL-based controller, where the line shows the average waiting time per training episode. Typically, the training curve exhibits a downward trend followed by convergence, as RL leverages accumulated experience and ultimately attains a local optimum. Moreover, we extract the average waiting time values from episodes 400 to 700 and compute their corresponding standard deviations. The findings reveal that CycLight and its variant, Advance CycLight, demonstrate standard deviations of 25.65 and 26.97, respectively. While single PDQN and MAADDPG exhibit standard deviations of 34.03 and 42.54, respectively. It is evident that both CycLight and Advance CycLight converges to the most optimal and stable policy with a narrow deviation range.

\begin{table*}[!ht]
\begin{center}
\caption{ATSC performance in the synthetic traffic grid.}
\label{tab3}
\begin{tabular}{c c c c c c c}
\hline
& \multirow{2}*{CycLight} & Single & \multirow{2}*{MAADDPG}  & \multirow{2}*{BackPressure} & Adaptive & Advance \\
&  & PDQN   & &   & Webster  & CycLight \\
\hline
Waiting time (s/veh) & {\textbf {205.15}} &	276.35 & 256.88 & 223.75 & 271.68 & {\textbf {205.31}}  \\
Throughput (veh) & {\textbf {4161}} &	3909 &	3987 &	4090 &	3815 &	{\textbf {4157}} \\
\hline
\end{tabular}
\end{center}
\end{table*}

\subsubsection{Evaluation results}
\label{S4_C_2}

\begin{figure*}[!ht]
  \centering
  \subfloat[]{\includegraphics[width=0.48\textwidth]{ 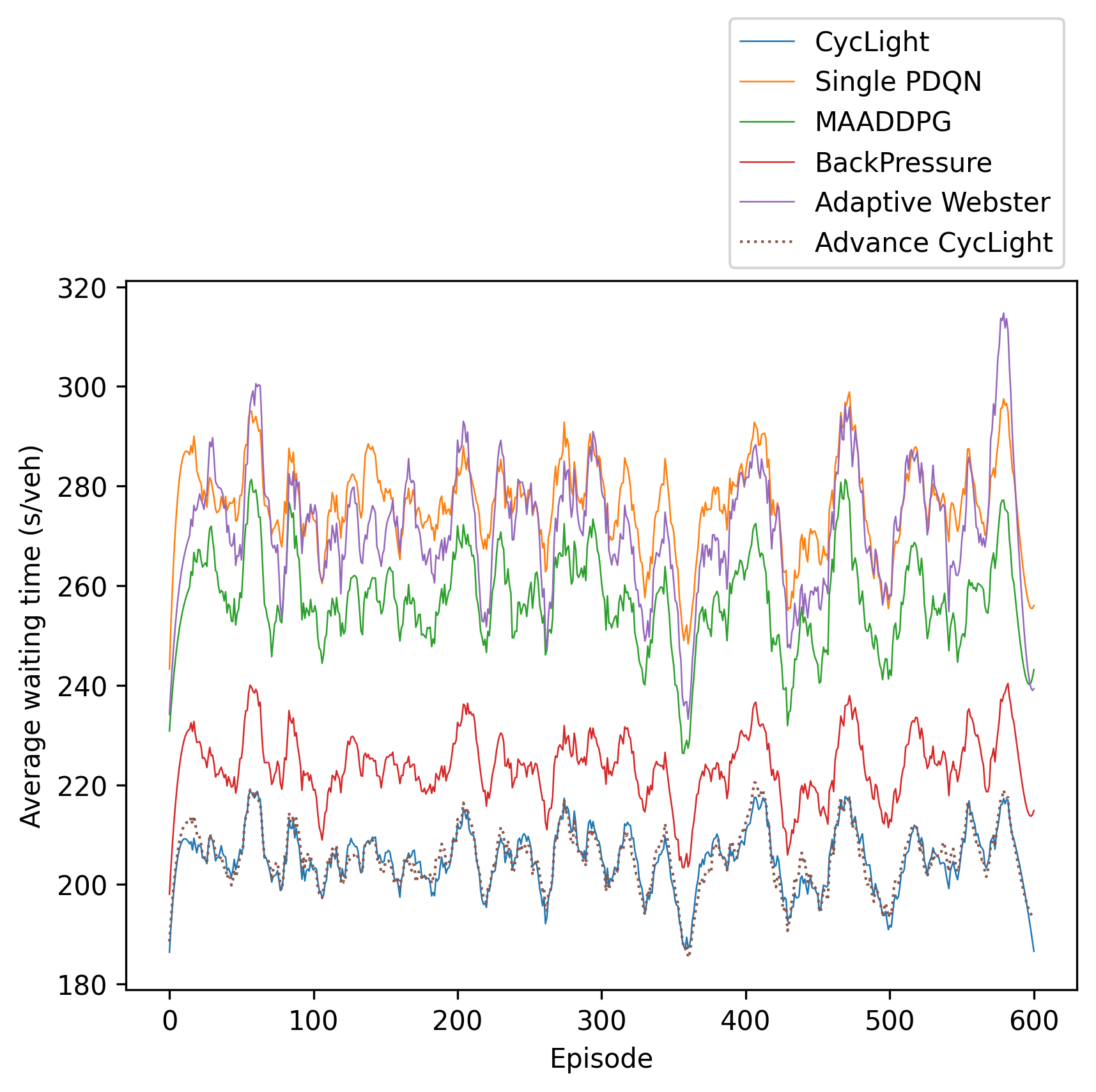}
  \label{F8_a}}
  \hfill
  \subfloat[]{\includegraphics[width=0.48\textwidth]{ 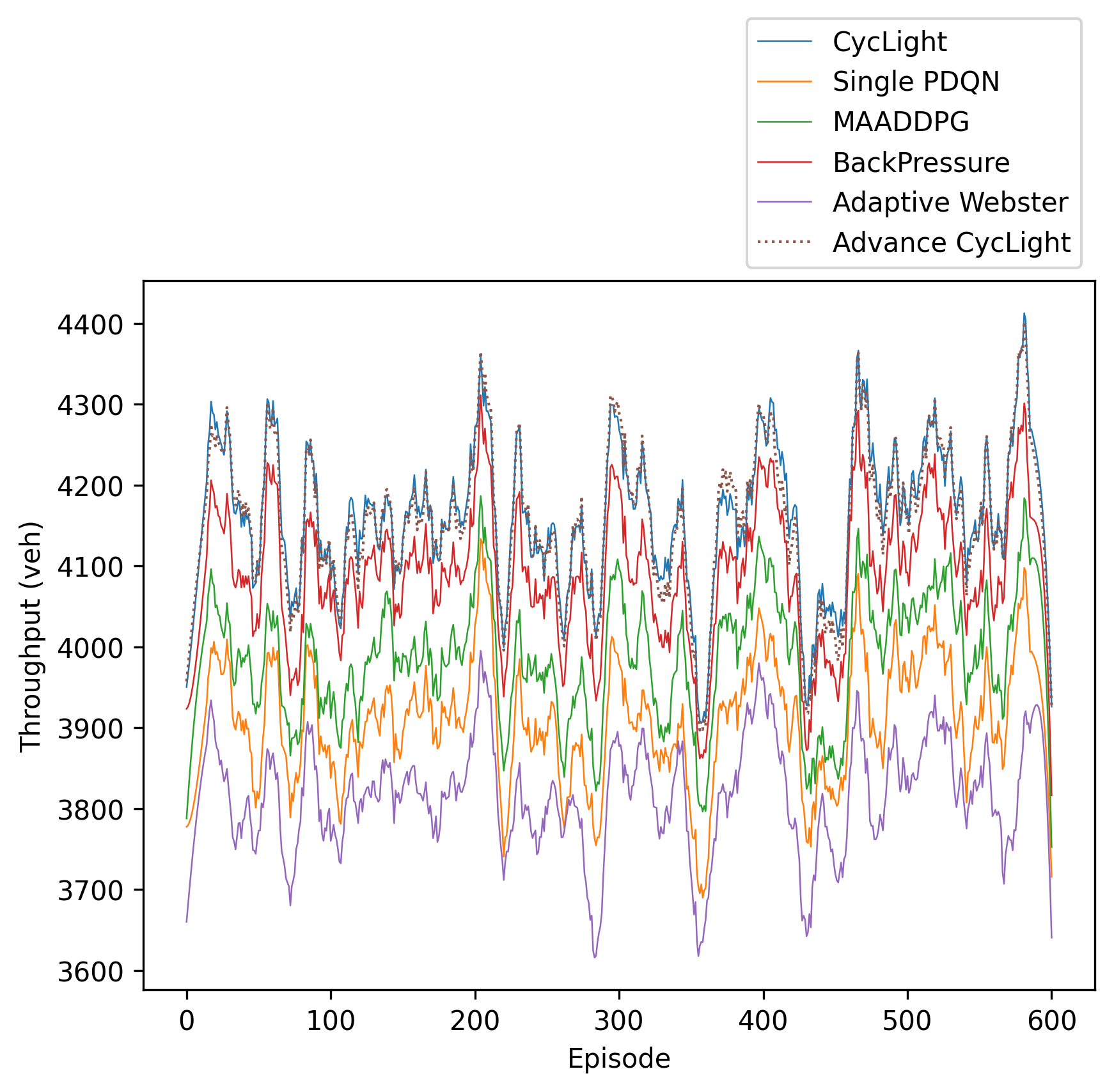}
  \label{F8_b}}
  \caption{Evaluation results over 600 episodes. (a) and (b) depict the average waiting time and throughput, respectively.}
  \label{F8}
  \end{figure*}

The well-trained and fine-tuned controllers are evaluated across 600 distinct episodes, each with a unique random seed by performing $\zeta \left( {E{\rm{ + }}1} \right){\rm{ = }}\zeta \left( E \right){\rm{ + }}1$. Various traffic flow patterns are simulated to assess the performance of different controllers. Figure \ref{F8} illustrates the average waiting time and the throughput of the whole network for each episode. Both CycLight and Advance CycLight excel in optimizing waiting time by maximizing throughput. Particularly, they outperform the suboptimal BackPressure method by an impressive margin of 8.31\% in terms of reducing average waiting time, as displayed in Table \ref{tab3}.

\begin{figure*}[!h]
  \centering
  \subfloat[]{\includegraphics[width=0.46\textwidth]{ 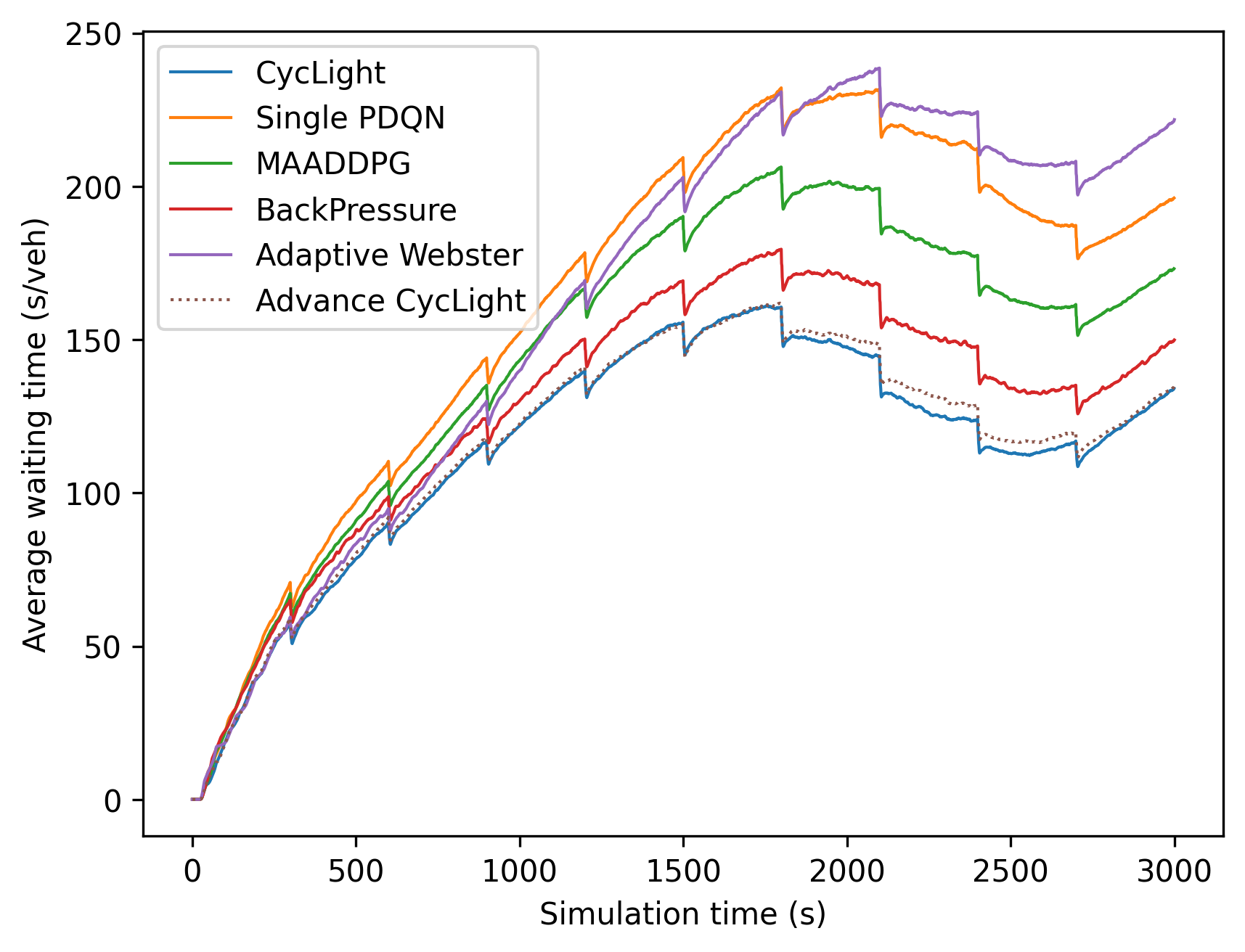}
  \label{F9_a}}
  \hfil
  \subfloat[]{\includegraphics[width=0.46\textwidth]{ 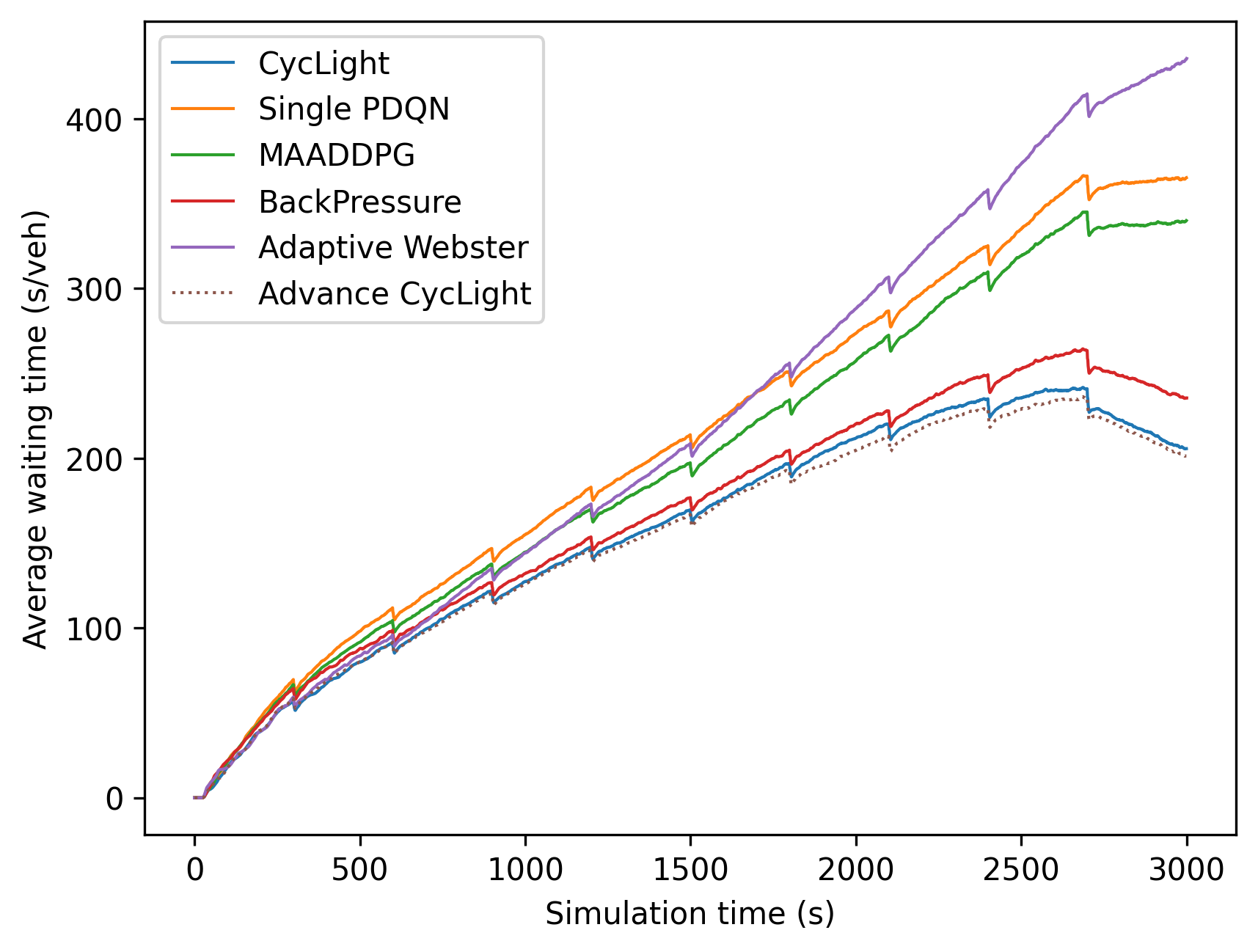}
  \label{F9_b}}
  \caption{Average waiting time within one episode. (a) for medium traffic demand, while (b) for high traffic demand. }
  \label{F9}
  \end{figure*}

Furthermore, we designed two representative traffic demands for further evaluation. Figure \ref{F9} illustrates the average waiting time over simulation time for the sampled episodes, with Figure \ref{F9} (a) and (b) representing medium and high traffic demand, respectively. As anticipated, both CycLight and Advance CycLight exhibit lower congestion levels and faster recovery. They effectively maintain low waiting times even during peak periods, while the other methods fall short. Encouragingly, the experimental results of {\bf Scenario B} demonstrate that our advanced control strategy in Advance CycLight does not compromise its performance. The robustness of our proposed method against information transmission delay is effectively showcased.

\subsubsection{Effect of attention mechanism}
\label{S4_C_3}

\begin{figure}[!ht]
  \centering
  \includegraphics[width=0.33\textwidth]{ 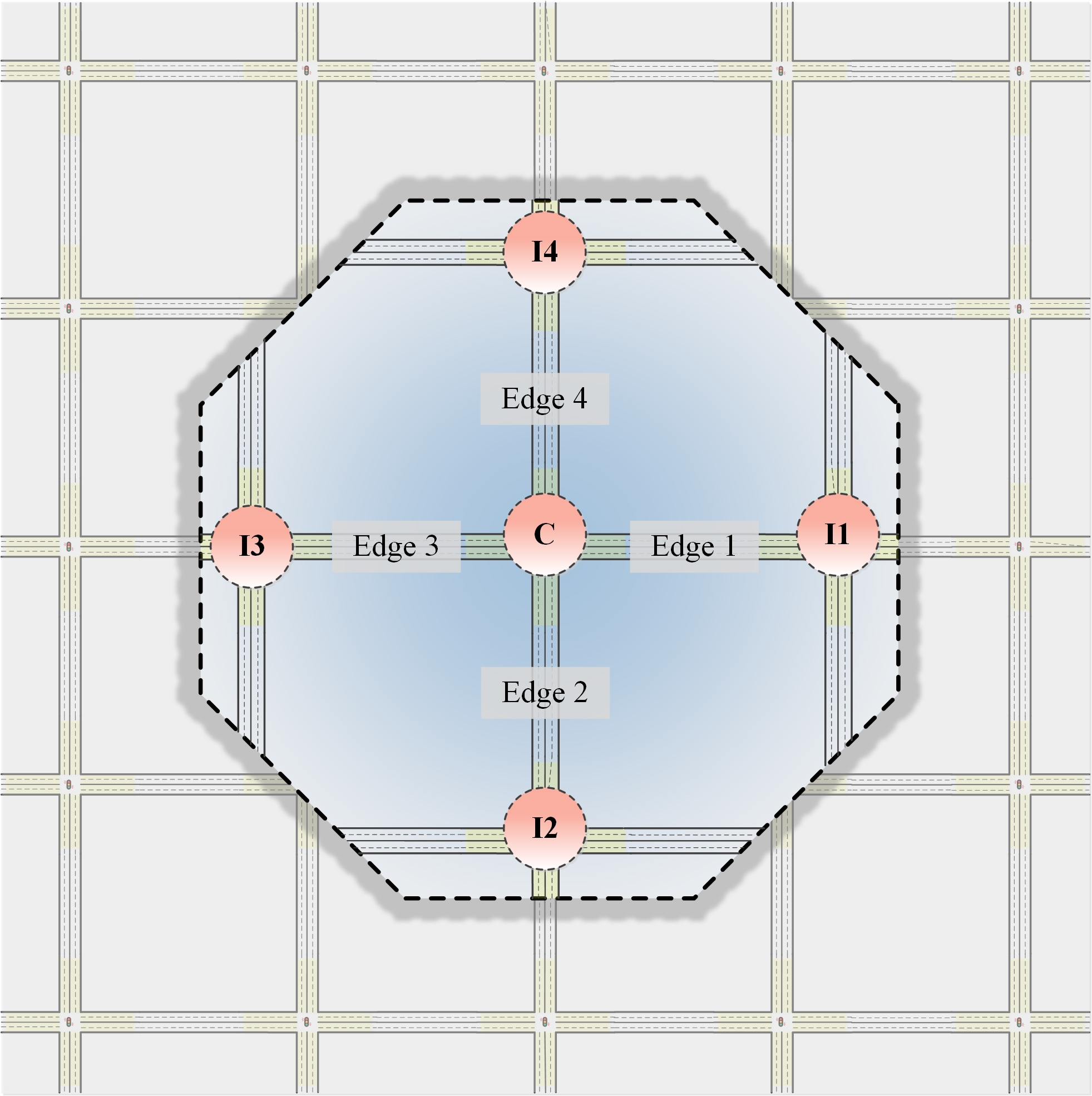}
  \caption{The selected intersection $C$ and its neighbors.}
  \label{F10} 
\end{figure}

\begin{figure}[!ht]
  \centering
  \includegraphics[width=0.43\textwidth]{ 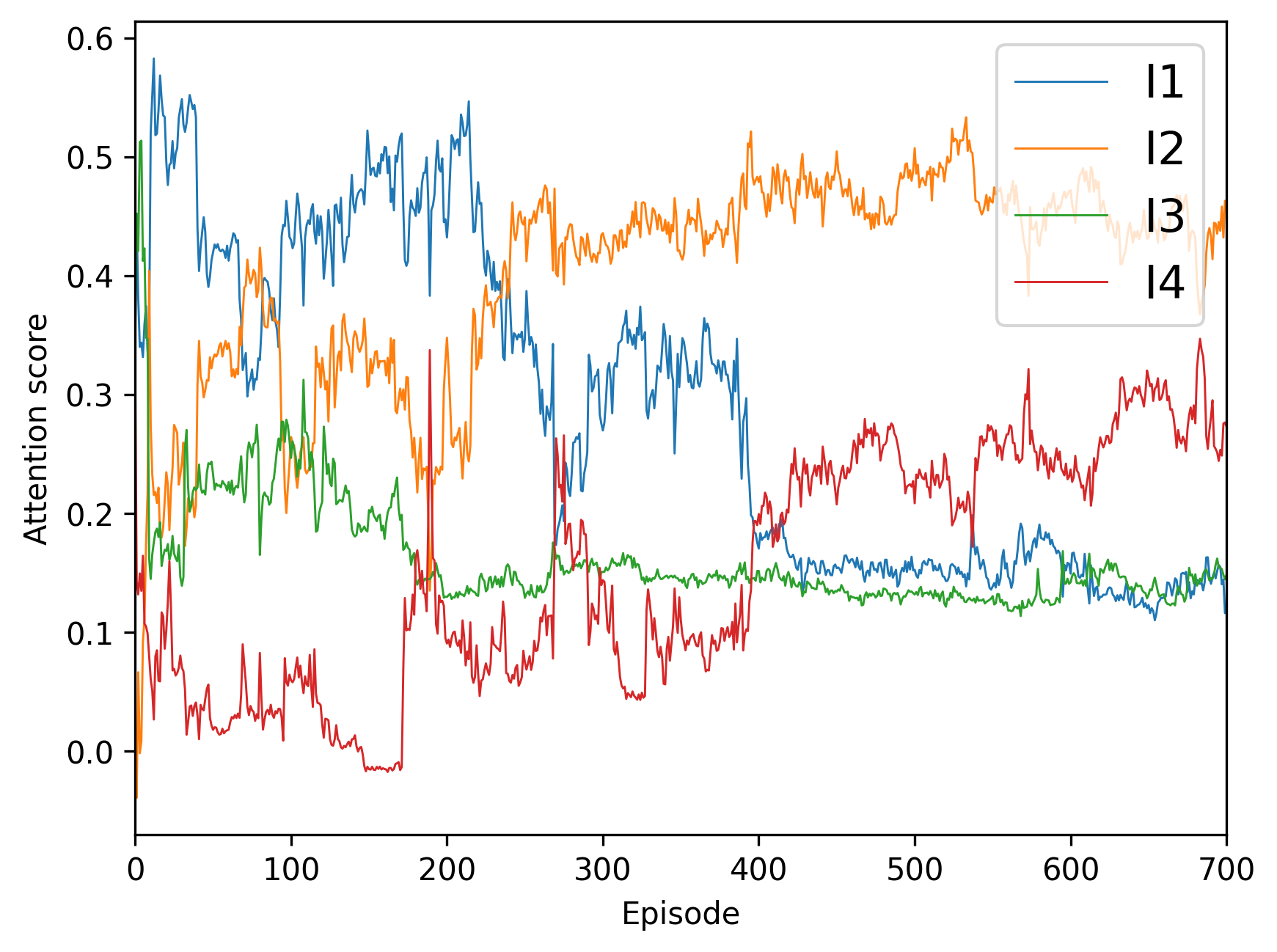}
  \caption{Attention scores of intersection $C$.}
  \label{F11}
  \vspace{-0.5cm}
\end{figure}

 To clarify how CycLight benefits from attention mechanism, we select an intersection $C$ located at the central traffic grid as the experimental target, observing the trend of its attention scores over the episodes. Accordingly, the neighbors $N_C = \{\rm{I1,I2,I3,I4}\}$ depicted in Figure \ref{F10} contribute their respective information to $C$ during the training process.  Given that the major traffic flows $F1$ and $F2$ surpass minor flows $f1$ and $f2$ (as shown in Table \ref{tab1}), it becomes evident that Edge 2 and 4 are more prone to severe traffic congestion compared to Edge 1 and 3. As expected, Figure \ref{F11} shows that intersection $\rm{I2}$ exerts the highest impact on $C$, followed by $\rm{I4}$. Consequently, the RL agent stationed at $C$ would dynamically adjust the cycle and splits to accommodate the major traffic flow in the north-south direction.


\section{Conclusions}
\label{S5}

In this paper, we propose a novel cycle-level RL-based approach, namely CycLight, aiming to enhance the traffic efficiency of the NATSC system. The proposed CycLight adopts the cycle-level TSC logic, leveraging PDQN agents to perform discrete-continuous hybrid actions. By jointly evolving the discrete action and continuous parameters, CycLight aims to identify the optimal cycle length while avoiding exhaustive search of continuous splits. Moreover, we establish a decentralized framework that promotes efficient and scalable cooperation among the agents. To account for the influence of surroundings on the current intersection, an attention mechanism is embedded in the approximating process of DNNs to adjust the weight accordingly. To the best of our knowledge, this is the first paper to adopt MARL with a discrete-continuous hybrid action space for cycle-level NATSC.

The proposed CycLight approach has been tested using SUMO. Experiments in a large-scale 5*5 traffic grid effectively substantiate the superiority, scalability, and robustness of our approach. Notably, the CycLight surpasses other state-of-the-art methods by exhibiting remarkable reductions in average waiting time and notable improvements in network throughput. Moreover, our proposed method displays commendable resilience against information transmission delays, as evidenced by the experimental findings derived from the advance control strategy.

For future work, given the flexibility of PAMDP settings offered by CycLight, incorporating pedestrian safety into RL training holds promise. Furthermore, a small-scale field experiment would better illustrate the practical superiority of CycLight, thus bridging the gap between theoretical advancements and real-world applications.

\section*{Acknowledgements}
This work was supported in part by the National Key Research and Development Program of China (No. 2022ZD011 5600); and in part by the National Natural Science Foundation of China (No.52072067)

\bibliographystyle{cas-model2-names}

\bibliography{CycLight.bib}

\end{document}